\documentclass[11pt]{article}
\usepackage{nodalida2021}
\usepackage{times}
\usepackage{url}
\usepackage{latexsym}
\usepackage{etaremune}
\usepackage{multirow}
\usepackage{graphicx}
\usepackage{longtable}
\usepackage{hyperref} % href
\usepackage{array}

\usepackage{xcolor}
\usepackage{siunitx}

\aclfinalcopy % Uncomment this line for the final submission
\title{Quantitative Evaluation of Alternative Translations in a Corpus of Highly Dissimilar Finnish Paraphrases}

\author{Li-Hsin Chang, Sampo Pyysalo, Jenna Kanerva, and Filip Ginter \\
  TurkuNLP Group \\
  Department of Computing \\
  Faculty of Technology \\
  University of Turku, Finland \\
  {\tt \{lhchan, sampyy, jmnybl, figint\}@utu.fi}}

\date{}

\begin{document}
\maketitle
\begin{abstract}
In this paper, we present a quantitative evaluation of differences between alternative translations in a large recently released Finnish paraphrase corpus focusing in particular on non-trivial variation in translation. We combine a series of automatic steps detecting systematic variation with manual analysis to reveal regularities and identify categories of translation differences. We find the paraphrase corpus to contain highly non-trivial translation variants difficult to recognize through automatic approaches. %All the tools and resources introduced in this study are made available under open licenses from [REDACTED].
\end{abstract}

\section{Introduction}

%Due to the lack of large corpora of manually identified alternative translations from multiple sources, analyses of variation in translation have largely focused on individual sources with multiple translations or noisy automatically aligned resources. \todo{confirm that true, add cites}
The study of translation language for Finnish has largely focused on individual linguistic features. 
The debate on the existence of translation universals sparked the well-developed research line of comparing translated and original language. Examples of such studies include the comparison of nonfinite structures in translated and original Finnish \citep{non-finite-construction, NFC-prose}, and investigation of subject changes in translations using a French-Finnish parallel corpus \cite{frfi-subject-change}.
% Suomessa  käännössuomen merkittävin  hanke  oli Joensuun  yliopiston  tutkimushanke,  jonka  yhteydessä  koottiin myös Käännössuomen   korpus.Tämä   korpus   sisältää   sekä   käännössuomen   että alkuperäissuomen   tekstejä   ja   mahdollistaa näiden   kahdenkielimuodonvertailun.Tutkimushanke on innoittanut myös useita pro gradu -töitä Suomen yliopistoissa. https://www.utupub.fi/bitstream/handle/10024/124757/JomppanenKristiinagradu2016.pdf?sequence=2&isAllowed=y
Variation in alternative translations is less studied.
\citet{retranslations} qualitatively compare the degree of domestication in language use in Finnish first translations and retranslations.
While this study is done qualitatively, several paraphrase corpora with translated language have been released more recently, enabling research from a quantitative prospective. Such corpora include Opusparcus \cite{CREUTZ18.131} and TaPaCo \cite{Scherrer2020TaPaCo}, both constructed automatically using language pivoting and containing Finnish subsets.

Recently, the Turku Paraphrase Corpus has become available \cite{kanerva2021finnish}, consisting of paraphrase pairs, of which the vast majority are manually selected from the OpenSubtitles\footnote{\url{http://www.opensubtitles.org}} dataset. The construction of the paraphrase corpus capitalizes on the fact that many movies and TV shows have multiple independently produced translations. The selection is carried out manually, comparing side-by-side the two lexically maximally distant subtitle versions for each movie or TV show and selecting instances of paraphrases. Upon selection, the candidate pairs are assigned to a category such as \emph{paraphrase in any context} or \emph{paraphrase in this context but not universally}, etc. The Turku paraphrase corpus is substantial in size, with 45,000 manually extracted, naturally occurring paraphrase pairs (a paraphrase pair henceforth refers to two segments of text, each about a sentence long or slightly longer), and a further 7,900 pairs created by editing an extracted pair so as to obtain a fully context-independent paraphrase.

Due to the way in which it was constructed, the corpus is directly applicable to the study of translation language and in particular to the analysis of variation in translation. The unique value of the corpus for this purpose is that it consists mostly of fully manually selected translation variants focused on lexically and structurally dissimilar pairs. These are very difficult to extract automatically: automatic methods can reliably identify only simple variation, while lexically and structurally substantially different pairs are very difficult to automatically distinguish from non-paraphrases, i.e.\ phrases that are not alternative translations.

% SMP: suggest to rephrase more generally: rather than saying that we're studying the corpus, I would suggest to say that we're using the corpus to study variation in translation.
In this paper, we will characterize the paraphrase corpus in terms of translation language, focusing especially on the types of variation (e.g. synonym usage, redundancy or verbosity) occurring in the data. Our aim is to establish whether the corpus can be of utility to translation language modelling and machine translation system evaluation. To this end, we will focus on two main questions: (a) how easily could the translation pairs be extracted automatically, and (b) what are the main types of variation exhibited by the pairs.

\section{Corpus statistics and pre-processing}

The full corpus includes 45,000 naturally occurring paraphrases and 7,900 pairs obtained by rewriting a previously extracted example. The source of these paraphrases is in the vast majority of cases alternative translations of subtitles, with a small section originating from news headings.
To construct a lexically and structurally diverse paraphrase corpus, the annotators were instructed to only select non-trivial paraphrase candidates, avoiding simple, uninteresting changes such as minor differences in inflection and word order.\footnote{Finnish has relatively free word order and reordering can be trivially detected automatically.} %As a result, the corpus consists of very few paraphrases that result from simple word reordering and/or variation of word form, despite Finnish being an inflectional language with a relatively loose word order.
For the analysis in this paper, we use the training section of the corpus, restricting further exclusively to examples originating from OpenSubtitles. This gives 34,561 naturally occurring paraphrase pairs and 5,445 rewritten paraphrases. Each naturally occurring paraphrase pair in the corpus have a numerical label manually assigned by an annotator from the following set:
\texttt{4}: universally paraphrase regardless of context, \texttt{3}: paraphrase in the given context but not universally, \texttt{2}: related but not paraphrase. Additionally, those annotated as \texttt{4} can be assigned one or several flags which sub-categorize different types of paraphrases: \texttt{>} or \texttt{<}: universal paraphrase in one direction but not the other, \texttt{s}: substantial difference in style, \texttt{i}: meaning-affecting difference restricted to a small number of morphosyntactic features. By contrast to the original paraphrases, the rewrites are always full, universally valid paraphrases, i.e. label \texttt{4}. The rewriting process strives to change as little of the original sentences as possible: these include simple fixes such as word or phrase deletion, addition or re-placement with a synonym or changing an inflection, while more complicated changes are avoided. The rewrites are thus an efficient way to obtain full paraphrases in terms of corpus creation.
The label distribution of the Turku paraphrase corpus subset used for later analysis is shown in Table~\ref{tab:label-dist}.

\begin{table}[t]
    \centering
    \begin{tabular}{ll}\hline
        Universal paraphrases & 14,986 \\
        \hspace{3mm} Label 4   & 8,578 \\
        \hspace{3mm} Label 4s  & 963 \\
        \hspace{3mm} Rewrites  & 5,445 \\\hline
        Context-dependent paraphrases  & 24,757       \\
        \hspace{3mm}(Label 3 or has \texttt{<}, \texttt{>}, or \texttt{i} flags)  &       \\\hline
        Related but not paraphrase  & 263       \\\hline
        \textbf{Total}     & \textbf{40,006}  \\\hline
        
      %  \textbf{Label} & 4 & 4$>$ & 4s & 4$>$s & 4i & 4$>$i & 4is & 4$>$is & 3 & 2 \\
    %  14,023  4+rew
    % 14,262 4> 
    %  963  4s
    % 832  4>s
    %  1,116 4i
    %  1,227 4>i
    %  99 4is
    % 60 4>is
    % 7,161 3
    % 263 2 not 437
    
    % All other: 24757 + 437 (label 2)  = 25,194                        
    \end{tabular}
    \caption{Label distribution of paraphrases from the subset of alternative subtitle translations in Turku paraphrase corpus training set.}
    %Context-dependent paraphrases include pairs assigned the numerical labels \texttt{3} or \texttt{4}, but are not \texttt{4} or \texttt{4s}. These include label \texttt{3} (paraphrase in the given context but not universally), as well as all paraphrases with flags \texttt{>}, \texttt{<} (universal in one direction but not the other) or \texttt{i} (meaning-affecting difference in morphosyntactic features).
    \label{tab:label-dist}
\end{table}

% original
%\begin{table*}[t]
%    \centering
%    \begin{tabular}{lllllllllll}\hline
%        \textbf{Label} & 4 & 4$>$ & 4s & 4$>$s & 4i & 4$>$i & 4is & 4$>$is & 3 & 2 \\
%        \textbf{Occurrences} & 14,023 & 14,262 & 963 & 832 & 1,116 & 1,227 & 99 & 60 & 7,161 & 437 \\\hline
%    \end{tabular}
%    \caption{Label distribution of the data. Label 4 includes the 5,445 rewrites.}
%    \label{tab:label-dist}
%\end{table*}

For the purpose of the subsequent analysis, we parse the paraphrases using the Turku Neural Parser Pipeline \cite{udst:turkunlp,kanerva2020lemmatizer}, a state-of-the-art parser producing POS and morphological tags, word lemmas, as well as dependency trees in the Universal Dependencies scheme \cite{nivre2016universal}. We use the model trained on UD\_Finnish-TDT v2.7 corpus, which utilizes the pre-trained FinBERT language model in tagging and dependency parsing \cite{virtanen2019multilingual}.\footnote{Model available at \url{https://turkunlp.org/Turku-neural-parser-pipeline/models.html}}

\section{Analysis of variation}

\subsection{Automatic categorization}

\begin{figure}
    \centering
    \includegraphics[width=0.5\textwidth]{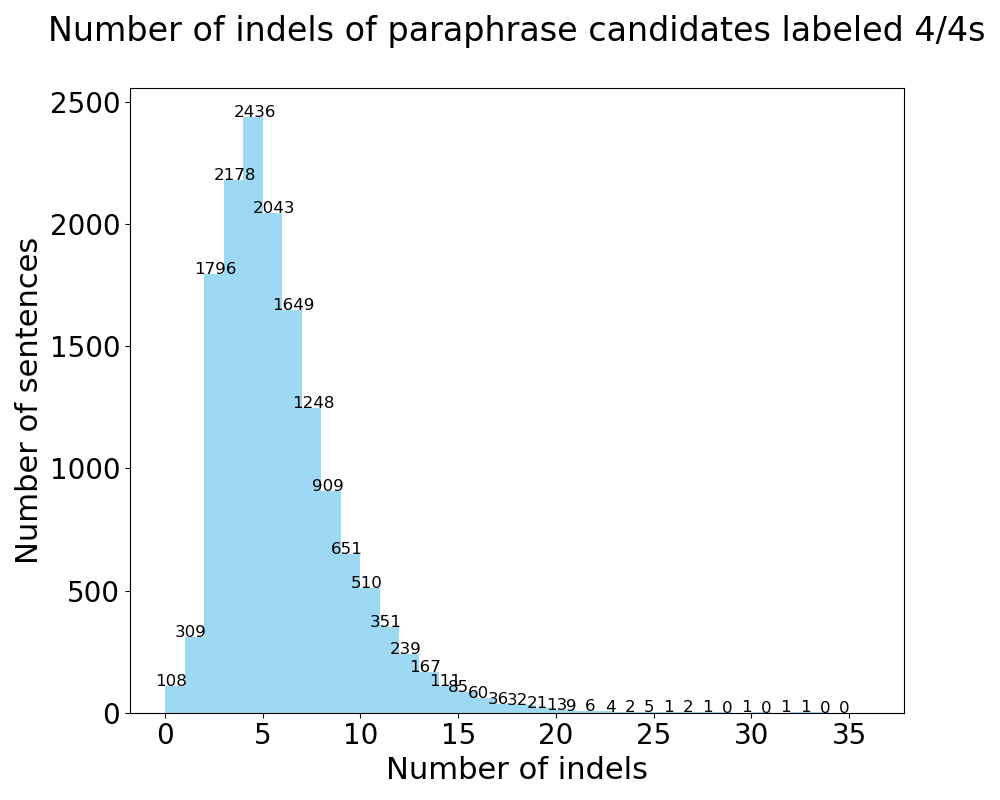}
    \caption{Distribution of the number of lemma indels for universal paraphrases labeled \texttt{4}/\texttt{4s} including rewrites.}
    \label{fig:lemma-indel-dist}
\end{figure}

To investigate and categorize the paraphrase pairs by the form of variation, we calculate the difference in the set of lemmas (i.e. insertions/deletions of lemma, henceforth lemma indels) for each pair, excluding punctuation characters from the analysis. Figure~\ref{fig:lemma-indel-dist} shows the distribution of the number of lemma indels for all universal paraphrases showed in Table~\ref{tab:label-dist} (paraphrases with labels \texttt{4} and \texttt{4s} including rewrites), i.e.\ all pairs equivalent in meaning regardless of their context. As a result of excluding trivial paraphrase candidates, less than 1\% (108 pairs) out of 14,986 pairs have zero lemma indels. Such pairs are formed purely by word reordering and/or changes in inflection.
We next investigate paraphrase pairs that can be accounted for by automatic synonym substitutions. We combine two resources to build a synonym dictionary for lemmas. The first resource is \texttt{Word2Vec} embeddings~\cite{word2vec2013} for lemmas trained from Suomi24 discussion fora texts\footnote{\url{dl.turkunlp.org/finnish-embeddings/finnish_s24_skgram_lemmas.bin}}. For each lemma, we take at most 15 closest lemmas in the vector space as synonyms using the \texttt{gensim} library \cite{gensim}. % vocab size 500,000
In addition, we supplement our synonym dictionary with Finnish WordNet \cite{linden14wnet} using the NLTK library \cite{nltk}. Out of the 14,878 pairs of paraphrases with lemma indels, 951 pairs ($\sim$6\%) have all of their lemma indels accounted by synonyms. An additional 7370 pairs ($\sim$49\%) have lemma indels partially accounted by synonyms.
The synonym dictionary only takes into account one-to-one synonyms. As a consequence, one-to-many synonyms and phrasal paraphrases are not included.

\begin{table}[!t]
\centering
\begin{tabular}{S[round-mode=places,round-precision=2] lll}
\textbf{Ratio} & \textbf{Word} & \textbf{Indel} & \textbf{Total} \\ \hline
0.447552447552448 & tosi (really) & 64 & 143 \\
0.408 & lakata (stop) & 51 & 125 \\
0.387323943661972 & ikävä (unfortunate) & 55 & 142 \\
0.384259259259259 & tahtoa (want) & 83 & 216 \\
0.370843989769821 & ihan (quite) & 145 & 391 \\
0.351398601398601 & todella (really) & 201 & 572 \\
0.344051446945338 & kai (perhaps) & 107 & 311 \\
0.341107871720117 & aivan (exactly) & 117 & 343 \\
0.339784946236559 & kyllä (truly) & 158 & 465 \\
0.339572192513369 & ikinä (never) & 127 & 374 \\
%0.338815789473684 & siis (therefore) & 103 & 304 \\
%0.331775700934579 & no (well) & 71 & 214 \\
%0.328320802005013 & varmasti (surely) & 131 & 399 \\
%0.317073170731707 & kuitenkin (anyway) & 52 & 164 \\
%0.316151202749141 & varmaan (probably) & 92 & 291 \\
%0.311804008908686 & täytyä (must) & 280 & 898 \\
%0.310408921933086 & taitaa (may) & 167 & 538 \\
%0.305343511450382 & laittaa (put) & 120 & 393 \\
%0.304511278195489 & panna ((to) place) & 81 & 266 \\
%0.302325581395349 & eräs (certain (someone)) & 65 & 215 \\
%0.297376093294461 & vaan (rather) & 102 & 343 \\
%0.292753623188406 & oikein (very) & 101 & 345 \\
%0.291866028708134 & työskennellä (work) & 61 & 209 \\
\end{tabular}
\caption{Most overrepresented words varying between different translations (minimum occurrence in corpus=50)}
\label{tab:indel-lemmas-by-ratio}
\end{table}

Table~\ref{tab:indel-lemmas-by-ratio} shows the lemmas that are most overrepresented  among the inserted or deleted words relative to their overall frequency. We find emphasizers (e.g.\ \emph{tosi (really)}), particles (e.g.\ \emph{kyllä (truly)}), auxiliary verbs, other functional words, and a small number of very common synonym pairs among the most frequently varying words.

To further focus on meaningful variation, we disregard all words with a dependency relation deemed functional in the Content-Word Labeled Attachment Score (CLAS) \cite{CLAS}, which is developed to evaluate dependency parsing with focus on content-bearing words.\footnote{These dependency relations are \texttt{aux} (auxiliary), \texttt{aux:pass} (passive auxiliary), \texttt{case} (pre/postposition), \texttt{cc} (coordinating conjunction), \texttt{clf} (classifier), \texttt{cop} (copula), \texttt{det} (determiner), \texttt{mark} (marker), \texttt{punct} (punctuation), \texttt{cc:preconj} (preconjunct), and \texttt{cop:own} (copula in possessive clauses).}
After disregarding these functional words, we are able to account for the variation in a further 82 paraphrase pairs. All of the above mentioned findings are summarized in Table~\ref{tab:categories}. 
As the variation in 13,608 pairs (i.e.\ full 90\% of the data) is not accountable by using the above automatic categories, we characterize these manually.

\begin{table}[t]
    \centering
    \begin{tabular}{ll}
        \textbf{4/4s} & \textbf{14986} \\\hline
        Word reordering & 1 \\
        Same lemma, same order & 27 \\
        Same lemma, different order & 80 \\
        CLAS & 82 \\
        Synonym & 945 \\
        Synonym + CLAS & 243 \\
        Others & 13608 \\ %Partial	0.44 (6544/14986), Others	0.47 (7064/14986)
    \end{tabular}
    \caption{Automatic classification of universal paraphrases labeled \texttt{4}/\texttt{4s} including rewrites.}
    \label{tab:categories}
\end{table}

\subsection{Manual categorization}

In the manual categorization, we sample 100 paraphrase pairs among those paraphrases where the variation is not fully explainable using the automatic metrics defined above. Each paraphrase pair is annotated in terms of 8 different variation categories: \emph{word-to-word}, \emph{word-to-phrase} and \emph{phrase-to-phrase} synonyms indicating a straightforward single word synonym replacement, a single word replaced with a synonymous phrase, or a phrase replaced with a synonymous phrase, \emph{redundancy or verbosity} for including additional words not strictly essential for the meaning, \emph{explicit pronouns} for explicitly including pronouns visible otherwise in the verb inflection, \emph{emphasizer} for including additional emphasis words (such as very), \emph{figurative language/idioms}, and \emph{uncertainty or hedging} where both statements express hedging with different markers.
% This "hedging" category is bit odd here, I suggest we drop it, or merge it with "emphasizers". -J

For each paraphrase pair a set of categories explaining the variation is annotated. In Table~\ref{tab:manual} we plot the frequency of each category, showing the straightforward single word synonym replacement being by far the most frequent category, occurring in 61\% of the paraphrase pairs. However, albeit word-to-word replacement being frequent, it rarely accounts for the whole variation in the pair. Only 12\% of the paraphrases include word-to-word synonyms as sole variation category, other instances occurring in combination with at least one additional variation category.

% note that these are annotated as sets, not lists (each category occurs at most once with each paraphrase)

\begin{table}[!t]
\centering
\begin{tabular}{lll}
\textbf{Category} & \textbf{Count} & \textbf{Ratio} \\ \hline
Word-to-word synonym       & 61 & 34\% \\
Word-to-phrase synonym     & 33 & 18\% \\
Phrase-to-phrase synonym   & 22 & 12\% \\
Redundancy or verbosity    & 21 & 12\% \\
Explicit pronouns          & 16 & 9\% \\
Emphasizers                & 14 & 8\% \\
Figurative language/idioms &  9 & 5\% \\
Uncertainty or hedging     &  3 & 2\% \\
\end{tabular}
\caption{Manual analysis results}
\label{tab:manual}
\end{table}

\subsection{Amount of Non-elementary Variation}

We measure the proportion of non-elementary variation in the alternative translations in terms of percentage of text (in terms of alphanumeric characters) in the manually extracted paraphrase pairs, out of the total amount of the source material that the annotators processed. The proportion is 15.8\%, meaning that approximately every sixth line was considered to be dissimilar in an interesting manner by the annotators, enough to be included in the paraphrase corpus. The remaining 84\% of the text is reported by the corpus creators to be for the most part elementary variation, text without correspondence in the other subtitle version, conflicting erroneous translations, and rarely pairs that are meaningless without deep understanding of their broader context.
% The distribution of documents with respect to the proportion of varied alternative translations is plotted in Figure~\ref{fig:coverage}, demonstrating that independently produced translations in the OpenSubtitles data only rarely show substantial amount of non-elementary variation.

%%%% FILIP: I think the figure is meaningless and can be deleted

%\begin{figure}
%    \centering
%    \includegraphics[width=0.5\textwidth]{figs-motra21/paraphrase_coverage.png}
%    \caption{The distribution of documents with respect to the proportion of non-elementary variation found from them (paraphrase coverage).}
%    \label{fig:coverage}
%\end{figure}

\subsection{Language pivoting}

To establish the proportion of the manually extracted paraphrase pairs that could be identified through their source text, as well as to establish the feasibility of automatically aligning the paraphrase pairs with their English source, we use the OpenSubtitles section of the OPUS machine translation dataset and identify those pairs in our dataset that have at least one common English source segment in the English--Finnish OpenSubtitles section of OPUS. We normalize both Finnish and English texts by lowercasing and dropping all non-alphanumeric characters so as to maximize the recall. 

Such language pivoting is a common technique for mining cases of translation variation. Language pivoting targets candidates, where the same source-language segment is translated into two different target-language segments, using a corpus of aligned bilingual document pairs. The candidates are typically further filtered by various means to remove spurious alignments and other pairs which are not equivalent in meaning, despite sharing the same aligned source-language segment.

We find that 2,136 pairs were matched, a mere 6\% of all categories of paraphrase in the corpus (barring rewrites). Full 94\% of the paraphrase pairs cannot be reached through simple language pivoting at least on the level of full segments. Further, while the average length of texts found through pivoting is 3.8 tokens, the average length of texts in the data is 8.4 tokens. The pivoting thus unsurprisingly  biases towards short segments, that are more likely to be appropriately aligned and identified. Clearly, in order to align the paraphrase pairs with their (mostly English) source, a manual annotation step will be necessary.

\section{Discussion, Conclusions and Future Work}

In this paper, we have presented a quantitative analysis of a large, manually extracted paraphrase dataset from the point of view of translation language, and especially its non-elementary variation. Our findings are two-fold. Firstly, we demonstrated that in the case of OpenSubtitles --- a very widely used corpus in machine translation --- the proportion of non-elementary variation in alternate translations is relatively small, at 16\% of the text. Secondly, we have shown that the paraphrase corpus contains highly non-trivial translation variants that are difficult to account for through simple heuristics and can thus serve for further study in translation language without biasing the results towards simpler examples that can be identified automatically.

The corpus in its current form can serve as a resource for evaluating robustness of different evaluation metrics. Quora Question Pairs (QQP)\footnote{\url{data.quora.com/First-Quora-Dataset-\\Release-Question-Pairs}} and the QQP subset of Paraphrase Adversaries from Word Scrambling (PAWS) \cite{zhang-etal-2019-paws} have been used to evaluate the robustness of machine translation and image captioning metrics \cite{bert-score}. QQP is a collection of question headings from the Quora forum labeled as either duplicate or not, while PAWS is an adversarial dataset automatically generated from QQP and Wikipedia to contain highly lexically similar paraphrases and non-paraphrases. Based on our findings, the Turku paraphrase corpus serves as an interesting resource to be used in a similar manner to evaluate metric robustness. An obvious direction for future work is to align, through a combination of heuristics and manual annotation, the paraphrase pairs with their English source. This would result in a test set suitable for evaluation of machine translation systems in terms of their rephrasing ability, as well as for research on MT system evaluation methodology in presence of substantial rephrasing.
% One potential limitation of this corpus is the quality of the translations. While some subtitles may be produced by professional translators, some are contributed by amateurs, and may inevitably contain non-standard translations.
% On the other hand, the paraphrases deemed interesting by the corpus creators will also contain alternative translations that do not necessarily score high on the common MT metrics.

%An obvious direction for future work is to align, through a combination of heuristics and manual annotation, the paraphrase pairs with their English source. This would result in a test set suitable for evaluation of machine translation systems in terms of their rephrasing ability, as well as for research on MT system evaluation methodology in presence of substantial rephrasing. 

\section*{Acknowledgments}
The research presented in this paper was partially supported by the European Language Grid project through its open call for pilot projects. The European Language Grid project has received funding from the European Union’s Horizon 2020 Research and Innovation programme under Grant Agreement no. 825627 (ELG). The research was also supported by the Academy of Finland and the DigiCampus project. Computational resources were provided by \textit{CSC — the Finnish IT Center for Science}. We thank Veronika Laippala for her advice from a linguistic point of view.

%Do not number the acknowledgment section. Do not include this section
%when submitting your paper for review.

\bibliographystyle{acl_natbib}
\bibliography{nodalida2021}

\newpage
\onecolumn
\appendix
\section{Example instances of manual analysis categories}
\label{sec:appendix}

\begin{table}[h!]
\centering
\begin{tabular}{c|l|l}
& \textbf{Translation$_1$} & \textbf{Translation$_2$} \\ \hline
\multirow{4}{*}{\rotatebox[origin=c]{90}{\parbox{4em}{\textbf{Word -\\word}}}}
& Vasta ammuttu & Ammuttu hiljattain \\
& Olen pistämättömän hygieeninen. & Olen moitteettoman hygieeninen. \\
& Etkö mennyt poliisin luo? & Et mennyt poliisin puheille? \\
& [...] on luultavasti uusi identiteetti. & [...] on varmasti uusi henkilöllisyys. \\ \hline

\multirow{4}{*}{\rotatebox[origin=c]{90}{\parbox{4em}{\textbf{Word -\\phrase}}}}
& Anteeksi odotus. & Anteeksi, että kesti. \\
& En edes osaa näytellä. & En edes tiedä miten näytellä. \\
& On niin paljon valinnanvaraa. & On niin paljon mistä valita. \\
& Useimmat teistä tietävät [...] & Suurin osa teistä tietää, [...] \\ \hline

\multirow{4}{*}{\rotatebox[origin=c]{90}{\parbox{4em}{\textbf{Phrase -\\phrase}}}}
& Andrew ehti ensin. & Andrew oli vain nopeampi. \\
& Iän myötä [...] & Mitä vanhemmaksi tulin, sitä [...] \\
& Miksi hän tekee niin? Etkö ole utelias? & Etkö halua tietää miksi hän tekee niin? \\
& kuuluuko seuralaisennekin tilin osakkaisiin? & Kuuluuko tili myös seuralaisellenne? \\ \hline

\multirow{4}{*}{\rotatebox[origin=c]{90}{\parbox{4.5em}{\textbf{Figurative}}}}
& Olen täysin hereillä, [...] & Olen pirteä kuin peipponen [...] \\ 
& Ole nyt vain hiljaa. & Pidä nyt vain pääsi kiinni. \\
%& Ystävyytemme on kaukana kuin valo miljoonan vuoden päässä. & Ystävyyden valo ei saavuta meitä miljoonaan vuoteen.
& Teitkö sen tasataksesi tilit? & Teitkö sen päästäksesi tasoihin? \\
& Tiedä häntä. & En minä tiedä.  \\ \hline
\multirow{4}{*}{\rotatebox[origin=c]{90}{\parbox{3em}{\textbf{Emph.}}}}
& Jopa runoja. & Runojakin. \\
& En tiennyt koko säännöstä. & En edes tiennyt säännöstä. \\
& Mitä täällä tapahtui?	& Mitä ihmettä täällä on tapahtunut? \\
& [...] näen asiat selvemmin. & [...] näen kaiken aina selvemmin. \\ \hline
%Voit saada naisen joka tottelee joka käskyäsi. mutta haluat naisen, jolla on mielipiteitä?. & Saat siis naisen, joka tottelee käskyjäsi mutta haluatkin naisen, jolla on mielipide? \\
\multirow{4}{*}{\rotatebox[origin=c]{90}{\parbox{4.5em}{\textbf{Verbosity/\\ redund.}}}}
& Voin kertoa teille, että [...] & Se mitä voin kertoa teille, on että [...] \\
& Se, ketä etsit, on kuollut! & Se ihminen jota etsit on kuollut! \\
& Mihin voin laittaa tämän? Pedille. & Minne voin laskea tämän? Voit laittaa sen sängylle. \\
& Hae ensiapupakkaus vessan kaapista. & Hae ensiapupakkaus. Se on vessan kaapissa. \\ \hline
\multirow{3}{*}{\rotatebox[origin=c]{90}{\parbox{3em}{\textbf{Hedge}}}}
& [...] herättävätkö ne liikaa huomiota. & [...] että ne saattavat kiinnittää liikaa huomiota. \\
& Vihaan [...] luultavasti ehkä enemmän [...] & Vihaan [...] ehkä enemmänkin [...] \\
& Lapset taisivat [...] & Näyttää siltä, että lapset [...] \\ \hline
\end{tabular}
\caption{Examples of manual analysis categories. English translations in Table \ref{tab:my_label_eng}.}
\label{tab:my_label}
\end{table}

\begin{table}[h!]
\centering
\begin{tabular}{c|l|l}
& \textbf{Translation$_1$} & \textbf{Translation$_2$} \\ \hline
\multirow{4}{*}{\rotatebox[origin=c]{90}{\parbox{4em}{\textbf{Word -\\word}}}}
& Recently shot & Just shot \\
& I am spotless clean & I am perfectly clean \\
& Didn't you approach the police? & Didn't you talk to the police? \\
& [...] is likely a new identity. & [...] is surely a new ID. \\ \hline

\multirow{4}{*}{\rotatebox[origin=c]{90}{\parbox{4em}{\textbf{Word -\\phrase}}}}
& Sorry the wait. & Sorry, that it took long. \\
& I can't even perform. & I don't even know how to perform. \\
& The choice is so varied. & The choice is very broad. \\
& Most of you know [...] & The biggest part of you know, [...] \\ \hline

\multirow{4}{*}{\rotatebox[origin=c]{90}{\parbox{4em}{\textbf{Phrase -\\phrase}}}}
& Andrew made it there first. & Andrew was simply faster. \\
& With age [...] & The older I became,  [...] \\
& Why is he doing so? Aren't you curious? & Don't you want to know why he is doing so? \\
& Does your colleague also & Does the stock belong also \\
& \hspace{7mm}belong among the stock holders? &  \hspace{7mm}to your colleague? \\ \hline

\multirow{4}{*}{\rotatebox[origin=c]{90}{\parbox{4.5em}{\textbf{Figurative}}}}
& I am fully awake, [...] & I'm astir as a bird [...] \\ 
& Be quiet now. & Keep your mouth shut. \\
%& Ystävyytemme on kaukana kuin valo miljoonan vuoden päässä. & Ystävyyden valo ei saavuta meitä miljoonaan vuoteen.
& Did you do it to even the score? & Did you do it to get equal? \\
& God knows. & I don't know..  \\ \hline
\multirow{4}{*}{\rotatebox[origin=c]{90}{\parbox{3em}{\textbf{Emph.}}}}
& Quite the poem. & A poem. \\
& I didn't know of the rule as such. & I really didn't know of the rule. \\
& What happened here?	& What on earth happened here? \\
& [...] you see things more clearly. & [...] you always see everything more clearly. \\ \hline
%Voit saada naisen joka tottelee joka käskyäsi. mutta haluat naisen, jolla on mielipiteitä?. & Saat siis naisen, joka tottelee käskyjäsi mutta haluatkin naisen, jolla on mielipide? \\
\multirow{4}{*}{\rotatebox[origin=c]{90}{\parbox{4.5em}{\textbf{Verbosity/\\ redund.}}}}
& I can tell you that [...] & What I can tell you is that [...] \\
& The one you are looking for is dead! & The person you are looking for is dead! \\
& Where can I put this? On the bed. & Where can I lay this down?\\
&                                   & \hspace{7mm}You can put it on the bed. \\
& Fetch the first aid kit & Fetch the first aid kit.\\
& \hspace{7mm}from the cupboard in the washroom &  \hspace{7mm}It is in a cupboard in the washroom. \\ \hline
\multirow{3}{*}{\rotatebox[origin=c]{90}{\parbox{3em}{\textbf{Hedge}}}}
& [...] do they attract too much attention. & [...] that they may attract too much attention. \\
& I hate [...] presumably maybe more [...] & I hate [...] maybe even more [...] \\
& The kids might [...] & It seems that the kids [...] \\ \hline
\end{tabular}
\caption{Examples of manual analysis categories, best-effort translation to English.}
\label{tab:my_label_eng}
\end{table}

\end{document}